# Knowledge transfer between bridges for drive-by monitoring using adversarial and multi-task learning

J. Liu[1], M. Bergés[2], J. Bielak[2] & H.Y. Noh[1]

[1] *Civil & Environmental Engineering, Stanford University, CA 94305, USA*

[2] *Civil & Environmental Engineering, Carnegie Mellon University, PA 15213, USA*

ABSTRACT: Monitoring bridge health using the vibrations of drive-by vehicles has various benefits, such as low cost and no need for direct installation or on-site maintenance of equipment on the bridge. However, many such approaches require labeled data from every bridge, which is expensive and time-consuming, if not impossible, to obtain. This is further exacerbated by having multiple diagnostic tasks, such as damage quantification and localization. One way to address this issue is to directly apply the supervised model trained for one bridge to other bridges, although this may significantly reduce the accuracy because of distribution mismatch between different bridges' data. To alleviate these problems, we introduce a transfer learning framework using domain-adversarial training and multi-task learning to detect, localize and quantify damage. Specifically, we train a deep network in an adversarial way to learn features that are 1) sensitive to damage and 2) invariant to different bridges. In addition, to improve the error propagation from one task to the next, our framework learns shared features for all the tasks using multi-task learning. We evaluate our framework using lab-scale experiments with two different bridges. On average, our framework achieves 94%, 97% and 84% accuracy for damage detection, localization and quantification, respectively, within one damage severity level.

## 1 INTRODUCTION

Bridges are key components of transportation infrastructure. However, one in eleven bridges in the U.S. has been reported to be structurally deficient (ASCE. 2017). The high cost and time usually required to inspect aging bridges desperately calls for advanced sensing and data analysis techniques.

The use of vibration signals collected from traveling vehicles to monitor bridges (i.e., drive-by monitoring) has recently become a viable and low-cost alternative to traditional bridge health monitoring approaches. In drive-by monitoring, the data is noisier because the measurement is indirect, and many types of noise are involved. Thus, signal processing and data analysis techniques play an important role reducing this measurement uncertainty. Data-driven approaches have recently begun using supervised and semi-supervised machine learning techniques to extract informative features from the vehicle acceleration signals, demonstrating successful structural damage diagnosis under certain conditions (Liu et al. 2019). Yet applying these approaches at scale is still challenging because 1) it is time-consuming, and impractical to obtain vehicle vibrations with corresponding bridge damage labels from every bridge; 2) directly applying supervised models trained on available data (i.e., a set of bridges with known damage labels) to other bridges may result in significantly worse performance because data distributions are different among different bridges with different characteristics. Furthermore, these challenges become more significant when we have multiple damage diagnostic tasks (e.g., damage detection, localization, and quantification) which require multiple damage label spaces.

To overcome the above challenges, we introduce a transfer learning framework, namely multi-task domain adversarial neural networks (MT-DANN), for bridge health monitoring using vehicle vibration responses. Our MT-DANN enables us to diagnose damage in multiple bridges without requiring labeled data from every bridge, by transferring the learned model from a bridge with labeled data (source domain) to another bridge without labels (target domain).

A key attribute of our MT-DANN framework is that it trains a neural network in an adversarial way to extract damage-sensitive features that are also domain-invariant from vehicle acceleration signals. Since the features are domain-invariant, they can detect, localize, and quantify damage across multiple bridges without the need of damage labels from the target domain. Moreover, our framework computes a shared feature representation using multi-task learning, which simultaneously optimizes multiple damage diagnostic tasks. This improves error propagation from one task to the next (e.g., from localization to quantification). We evaluate our framework on lab-scale experiments with two different bridges. Our results show that our framework outperforms two baseline methods which do not apply domain adversarial training or multi-task learning.

## 2 DOMAIN ADAPTATION FORMULATION FOR DRIVE-BY MONITORING

In this section, we introduce the domain adversarial neural networks (DANN) which is the basic model of our framework.

We consider classification tasks that label vehicle acceleration signals as belonging to different classes of bridge damage. We denote bridges where labelled data are available as 'Source Domain' $D_s = \{(x_i^s, l_i^s, q_i^s)\}_{i=1}^{n_s}$; bridges without labelled data as 'Target Domain' $D_t = \{x_i^t\}_{i=1}^{n_t}$, where $x_i^s$ and $x_i^t$ are vehicle accelerations in the source and target domains, respectively; $l_i^s$ and $q_i^s$ are damage location and severity (labels) of source domain data, respectively; $n_s$ and $n_t$ are numbers of samples in the source and target domains, respectively. Note that the target domain data do not include labels, $\{l_i^t, q_i^t\}_{i=1}^{n_t}$. Our goal, and the goal of the domain adaptation, is to build a classifier to minimize the classification error of the target domain without using information about the target domain labels.

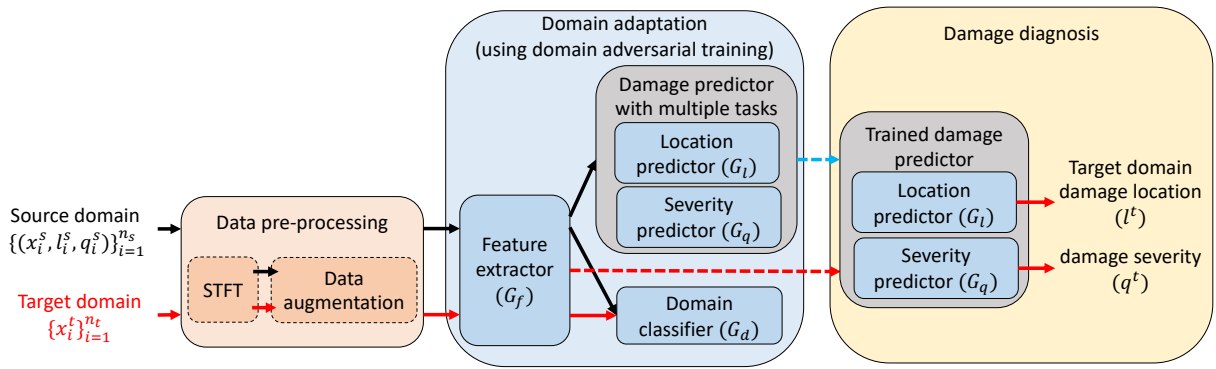

Figure 2. Our domain adaptation framework for vehicle-vibration-based bridge health monitoring. Black arrows represent labeled dataflow of the source domain, red arrows represent unlabeled dataflow of the target domain, and blue arrow represents the use of trained damage predictor. Dashed arrows represent the testing phase.

To better formulate our problem, there are two important physical facts of the vehicle-bridge interaction system (VBI) that need to be incorporated:
a) Vehicle acceleration is a high-dimensional signal that has a non-linear relationship with bridge properties and damage parameters. Thus, it is difficult to infer damage states for different bridges by directly inspecting the raw vehicle signals.
b) If we localize and quantify the damage sequentially, the uncertainty of damage magnitude estimation will contain powers of sinusoid functions in the denominators that result in a large propagation of error from one estimation to the next (Liu et al. 2019).

As a result, our approach should incorporate the physical insights of the VBI system, be able to extract damage-sensitive and domain-invariant feature representation, and be able to localize and quantify damage simultaneously using vehicle acceleration signals.

### 2.1 Domain adversarial training of neural networks

Domain adaptation, which learns a model from a source data distribution and generalizes the model to a different but related target data, consists of two main strategies, instance weighting adaptation and feature adaptation. The former reweighs samples to adapt domains that have different data distributions caused by sample selection bias or covariate shift. The latter aims to find the common feature representation of data from different domains using linear and non-linear feature extractor methods (Zhang, L. 2019.).

In our damage diagnosis framework, we use a feature adaptation approach, domain adversarial neural networks (DANN) (Ganin, Y. et al. 2016). DANN consists of three components, feature extractor, source predictor and domain classifier. The domain classifier is trained to distinguish the source domain data from the target domain data, and the feature extractor is trained simultaneously to confuse the domain classifier. A domain-invariant and class-discriminative feature representation is extracted from the input data by minimizing the source classifier loss and simultaneously maximizing the domain classifier loss. DANN is appealing because the minimax objective can be achieved by applying a gradient reversal layer (GRL), which can be easily incorporated into any existing neural network architectures that handle high-dimensional signals.

## 3 ADVERSARIAL KNOWLEDGE TRANSFER FRAMEWORK FOR DRIVE-BY MONITORING

In this section, we introduce our Multi-task domain adversarial neural network (MT-DANN), a novel damage diagnosis framework for drive-by monitoring. As shown in Figure 2, our framework contains three modules: (1) a data pre-processing module, (2) a domain adaptation module, and (3) a damage diagnosis module.

**Data pre-processing module:** In this module, we prepare the input data for MT-DANN by computing the time-frequency domain representation and conducting data augmentation. We first compute the short-time Fourier transform (STFT) of each vertical acceleration record of the vehicle traveling over the bridge to preserve the time-frequency domain information. Then, we arrange the signals to create the input for source and target domain. The size of input data is C×W×H, where C is the number of sensor channels on the vehicle. W and H are respectively the number of time segments and the sample frequencies of the STFT representation. To avoid over-fitting, data biases, and to provide sufficient information of each damage state, we conducted data augmentation on raw signals by adding white noise and randomly erasing samples. This data augmentation improves the robustness of the learned model.

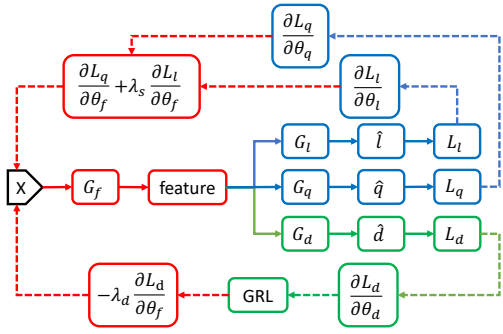

Figure 3. The architecture of MT-DANN. Solid and dash arrows represent forward computation and back-propagation, respectively. The red, blue and green parts show the feature extractor, damage predictor and domain classifier, respectively.

**Domain adaptation module:** In this module, we train our MT-DANN to obtain 1) a feature extractor that extracts domain-invariant and damage-sensitive features, and 2) a damage predictor that uses the features to predict damage states for both source and target domain data. This is achieved by jointly optimizing four components: feature extractor ($G_f$), location predictor ($G_l$), severity predictor ($G_q$) and domain classifier ($G_d$). This algorithm is implemented in a novel architecture shown in Figure 3. The parameters $\theta_f$ of the feature extractor $G_f$ are learned by maximizing the loss of the domain classifier, while the parameters $\theta_d$ of domain classifier $G_d$ are learned by minimizing the loss of the domain classifier. Also, the parameters $\theta_l$ and $\theta_q$ of location and quantification predictors are learned by minimizing the loss of location and severity predictors $G_l$ and $G_q$ to guarantee low source domain classification errors. The objective of MT-DANN is thus:

$$L(\theta_f, \theta_l, \theta_q, \theta_d) = \frac{1}{n_s} \sum_{x_i \in D_s} [L_l(G_l(G_f(x_i; \theta_f); \theta_l), l_i) \\ + \lambda_s L_q(G_q(G_f(x_i; \theta_f); \theta_q), q_i)] \\ - \frac{\lambda_d}{n} \sum_{x_i \in D_s \cup D_t} L_d(G_d(G_f(x_i; \theta_f); \theta_d), d_i),$$
(3)

where $L_l$, $L_q$, and $L_d$ are objective functions for the damage location and severity predictors, and the domain classifier; $d_i$ is the domain label of input signal $x_i$; $\lambda_s$ and $\lambda_d$ are hyper-parameters to trade off the three loss functions; $n = n_s + n_t$ is the total number of source and target domain data. To obtain parameters for each network, we solve the following optimization problems:

$$(\hat{\theta}_f, \hat{\theta}_l, \hat{\theta}_q) = \underset{\theta_f, \theta_l, \theta_q}{\arg\min} L(\theta_f, \theta_l, \theta_q, \theta_d),$$
(4)

$$\hat{\theta}_d = \underset{\theta_d}{\arg\max} L(\theta_f, \theta_l, \theta_q, \theta_d),$$
(5)

Importantly, based on physical insights of the VBI system, our novel approach simultaneously optimizes the two damage diagnostic tasks using multi-task learning. It computes shared features for these two tasks in order to reduce the error propagating from one task to the next. Detailed architectures of our networks are shown in Table 1. The justifications for choosing layers, kernel size, activation functions, etc. are based on empirical validations.

Table 1. The detailed architecture of each network.

| Network | Layer | Patch size | Input size | Activation |
|---|---|---|---|---|
| Feature extractor ($G_f$) | Conv2d | $64 \times 5 \times 5$ | $4 \times 64 \times 64$ | ReLU |
| | MaxPool | $2 \times 2$ | $64 \times 60 \times 60$ | |
| | Conv2d | $50 \times 5 \times 5$ | $64 \times 30 \times 30$ | ReLU |
| | MaxPool | $2 \times 2$ | $50 \times 26 \times 26$ | |
| | Conv2d | $50 \times 3 \times 3$ | $50 \times 13 \times 13$ | ReLU |
| Location predictor ($G_l$) | Flatten | | $50 \times 5 \times 5$ | |
| | Linear | $1250 \times 100$ | $1250 \times 1$ | ReLU |
| | Linear | $100 \times 4$ | $100 \times 1$ | Softmax |
| Severity predictor ($G_q$) | Flatten | | $50 \times 5 \times 5$ | |
| | Linear | $1250 \times 100$ | $1250 \times 1$ | ReLU |
| | Linear | $100 \times 5$ | $100 \times 1$ | Softmax |
| Domain Classifier ($G_d$) | Flatten | | $50 \times 5 \times 5$ | |
| | Linear | $1250 \times 100$ | $1250 \times 1$ | ReLU |
| | Linear | $100 \times 2$ | $100 \times 1$ | Softmax |

**Damage diagnosis module:** Finally, we input the domain-invariant feature extracted from target domain data using the feature extractor $G_f$ to the trained damage predictors, $G_l$ and $G_q$, to predict damage locations and severity levels of the target bridge. Also, the determination of the state of a bridge is made possible by introducing an additional label within the damage predictor whereby if there is no indication of detected damages, the bridge is considered healthy. For example, we use label value zeros to represent healthy states in the location predictor.

## 4 EVALUATION

In this section, we evaluate our framework on lab-scale experiments conducted on two structurally different bridges using three differently weighted vehicles.

**Experimental setup and the dataset:** A lab-scale vehicle-bridge interaction (VBI) system, as shown in Figure 4, was employed to collect the dataset. There are two 2.44-meter bridges (B #1 and B #2) with different dominant frequencies of 7.7 Hz and 5.9 Hz, and damping ratios of 0.13 and 0.07, respectively; three vehicles (V #1, V #2 and V #3) with different weights of 4.8 kg, 5.3 kg, and 5.7 kg, respectively, were driven over the bridge. Vertical acceleration signals were collected from four accelerometers placed on each vehicle (front chassis, back chassis, front wheel and back wheel) while they moved individually across the bridge at a constant speed (0.75 m/s). The sampling rate of the sensors is 1600 Hz.

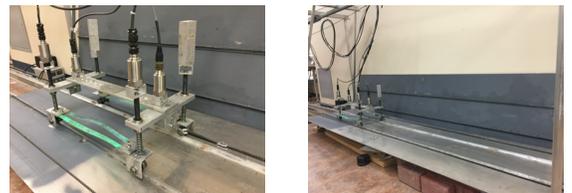

(a) Vehicle.  (b) Bridge

Figure 4: (a) the instrumented vehicle moving at controlled speeds and (b) the bridge that the vehicle passes.

Damage proxy is introduced by adding mass at different locations of the bridge. The magnitude of the attached mass ranges from 0.23 kg to 0.91 kg with an interval of 0.23 kg. A heavier mass means more severe damage since it appears as a more significant change from the initial condition (i.e., healthy state). For each damage severity level, the experiment is run with three different damage locations ($l$ is every quarter of the bridge span), and for each damage scenario, the experiment is run thirty times. In total, the dataset has 2 (bridges) × 3 (vehicles) × [3 (damage locations) × 4 (damage quantifications) + 1 (undamaged case)] × 30 (iterations) = 2340 (trials), and 2340 (trials) × 4 (sensors) = 9360 (records). Details of the experimental instrumentation can be found in (Liu et al. 2020.)

**Results and discussion:** We evaluate the performance of our MT-DANN framework on the lab-scale dataset for knowledge transfer between bridges. We have four damage diagnostic tasks, including binary damage detection, three-class damage localization, four-class damage quantification, and damage quantification within one damage severity level. For each damage diagnostic task, two knowledge transfers, from B #1 to B #2 and B #2 to B #1, using signals collected from three vehicles, were conducted on each vehicle, making a total of six evaluations. In addition, for each evaluation, we compare the performance of our framework with two baseline methods: MT-CNN and 2-step DANN. MT-CNN trains a multi-task convolutional neural network (having the same architectures as the feature extractor, location predictor, and severity predictor) on the source domain data and predicts on the target domain data without domain adaptation. 2-step DANN first uses a DANN to train on the source domain data and predict damage location on the target domain data, and in the second step, it uses another DANN to predict damage severity using the data with the same damage location predictions.

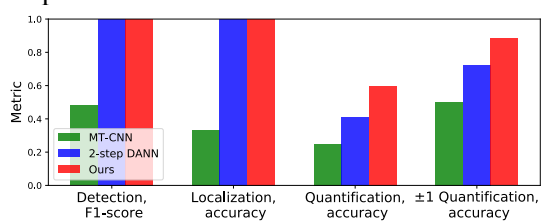

Figure 5. Damage detection, localization quantification results of two baseline methods and our framework for knowledge transfer from B #1 to B #2 using V #2 data.

Figure 5 presents the results for knowledge transfer from B #1 to B #2 using V #2 data. Our framework shows the best results. Overall, for all six evaluations, our framework is more than twice as accurate compared to MT-CNN on damage detection and localization tasks, and around 1.7 times more accurate on damage quantification tasks. Compared to 2-step DANN, our framework improves the damage quantification accuracy by 7 % on average, which provides evidence that simultaneously localizing and quantifying the damage reduces error propagation. We also observe that the accuracy of damage quantification is worse than the other tasks, because the damage severity in our experiments changes gradually.

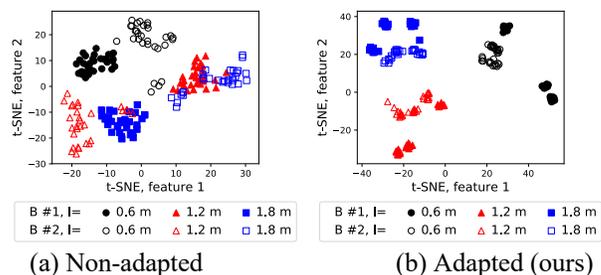

(a) Non-adapted      (b) Adapted (ours)

Figure 7. t-SNE (Maaten, L.V.D. & Hinton, G. 2008.) visualizations of the feature maps (a) without domain adaptation and (a) with domain adaptation on the damage localization task for the evaluation of B#1 → B #2 using V #2 data.

In addition, in Figure 7, we visualize the t-SNE embeddings (Maaten, L.V.D. & Hinton, G. 2008.) of the feature by MT-CNN and MT-DANN on the damage localization task for the B #1 → B #2 evaluation using V #2 data. We observe that the adaptation in our method makes the two distributions of features extracted from two bridges' data much closer. Compared to the baseline method without domain adaptation, our framework can discriminate different damage locations in both source and target domains.

## 5 CONCLUSIONS AND FUTURE WORK

We presented MT-DANN, a transfer learning framework that uses vehicle acceleration responses to diagnose structural damage of multiple bridges without requiring labeled data from every bridge. Our framework uses 1) DANN to learn a feature that is sensitive to damage and invariant to different bridges, and 2) multi-task learning to simultaneously optimize multiple damage diagnostic tasks for reducing error propagation from one task to the next. We evaluated our framework on lab-scale experiments with two structurally different bridge models and three differently weighted vehicles. Our framework achieves respectively 94%, 97% and 84% accuracy for detecting, localizing and quantifying damage of the target bridge. It outperforms two baseline approaches that do not apply domain adversarial training or multi-task learning.

Some near-future tasks are to modify our approach and use strong classifiers to improve damage quantification accuracy and test the scalability of our framework on both lab-scale and real-world vehicle-bridge systems.

REFERENCES

ASCE. 2017. ASCE's 2017 infrastructure report card.
Ganin, Y., Ustinova, E., Ajakan, H., Germain, P., Larochelle, H., Laviolette, F., Marchand, M. and Lempitsky, V. 2016. Domain-adversarial training of neural networks. *The Journal of Machine Learning Research*, 17(1), pp.2096-2030.
Liu, J., Bergés, M., Bielak, J., Garrett, J.H., Kovačević, J. and Noh, H.Y. 2019. May. A damage localization and quantification algorithm for indirect structural health monitoring of bridges using multi-task learning. *In AIP Conference Proceedings (Vol. 2102, No. 1, p. 090003).* AIP Publishing LLC.
Maaten, L.V.D. & Hinton, G. 2008. Visualizing data using t-SNE. *Journal of machine learning research*, 9(Nov), pp.2579-2605.
Zhang, L. 2019. Transfer adaptation learning: A decade survey. *arXiv preprint arXiv:1903.04687*.